\def\year{2020}\relax
\documentclass[letterpaper]{article} 
\usepackage{aaai20}  
\usepackage{times}  
\usepackage{helvet} 
\usepackage{courier}  
\usepackage[hyphens]{url}  
\usepackage{graphicx} 
\usepackage{graphicx}  
\usepackage{subcaption}
\newcommand{\Tstrut}{\rule{0pt}{2.3ex}}
\newcommand{\br}[1]{\underline{\textbf{#1}}}
\newcommand{\bb}{\textbf}
\newcommand{\citet}[1]{\citeauthor{#1} \shortcite{#1}}
\newcommand{\citep}{\cite}

\title{Why Attention? Analyze BiLSTM Deficiency and Its Remedies in the Case of NER}

\author{
Peng-Hsuan Li,\textsuperscript{\rm 1} Tsu-Jui Fu,\textsuperscript{\rm 2} Wei-Yun Ma\textsuperscript{\rm 1} \\
\textsuperscript{\rm 1}Academia Sinica, \textsuperscript{\rm 2}UC Santa Barbara \\
jacobvsdanniel@iis.sinica.edu.tw, tsu-juifu@ucsb.edu, ma@iis.sinica.edu.tw
}

\begin{document}

\maketitle

\begin{abstract}
BiLSTM has been prevalently used as a core module for NER in a sequence-labeling setup. State-of-the-art approaches use BiLSTM with additional resources such as gazetteers, language-modeling, or multi-task supervision to further improve NER. This paper instead takes a step back and focuses on analyzing problems of BiLSTM itself and how exactly self-attention can bring improvements. We formally show the limitation of (CRF-)BiLSTM in modeling cross-context patterns for each word -- the XOR limitation. Then, we show that two types of simple cross-structures -- self-attention and Cross-BiLSTM -- can effectively remedy the problem. We test the practical impacts of the deficiency on real-world NER datasets, OntoNotes 5.0 and WNUT 2017, with clear and consistent improvements over the baseline, up to 8.7\% on some of the multi-token entity mentions. We give in-depth analyses of the improvements across several aspects of NER, especially the identification of multi-token mentions. This study should lay a sound foundation for future improvements on sequence-labeling NER\footnote{Source codes: \url{https://github.com/jacobvsdanniel/cross-ner}}.
\end{abstract}

\section{Introduction}
\label{sec:introduction}

Named Entity Recognition (NER) is a core task for information extraction. Originally a structured prediction task, NER has since been formulated as a task of sequential token labeling. BiLSTM-CNN uses a CNN to encode each word and then uses bi-directional LSTMs to encode past and future context respectively at each time step. With state-of-the-art empirical results, most regard it as a robust core module for sequence-labeling NER~\citep{Ma:ACL2016,Chiu:TACL2016,Aguilar:NAACL2018,Akbik:COLING2018,Clark:EMNLP2018}.

However, each direction of BiLSTM only sees and encodes half of a sequence at each time step. For each token, the forward LSTM only encodes past context; the backward LSTM only encodes future context. For computing sentence representations for tasks such as sentence classification and machine translation, this is not a problem, as only the rightmost hidden state of the forward LSTM and only the leftmost hidden state of the backward LSTM are used, and each of the endpoint hidden states sees and encodes the whole sentence. For computing sentence representations for sequence-labeling tasks such as NER, however, this becomes a limitation, as each token uses its own midpoint hidden states, which do not model the patterns that happen to cross past and future at this specific time step.

This paper explores two types of cross-structures to help cope with the problem: Cross-BiLSTM-CNN and Att-BiLSTM-CNN. Previous studies have tried to stack multiple LSTMs for sequence-labeling NER~\citep{Chiu:TACL2016}. As they follow the trend of stacking forward and backward LSTMs independently, the Baseline-BiLSTM-CNN is only able to learn higher-level representations of past or future per se. Instead, Cross-BiLSTM-CNN, which interleaves every layer of the two directions, models cross-context in an additive manner by learning higher-level representations of the whole context of each token. On the other hand, Att-BiLSTM-CNN models cross-context in a multiplicative manner by capturing the interaction between past and future with a dot-product self-attentive mechanism~\citep{Conneau:EMNLP2017,Lin:ICLR2017}.

Section~\ref{sec:model} formulates the three Baseline, Cross, and Att-BiLSTM-CNN models, with Section~\ref{sec:xor},~\ref{sec:xor_crf} giving formal proof that patterns forming an XOR cannot be modeled by (CRF-)BiLSTM-CNN used in all previous work. Cross-BiLSTM-CNN and Att-BiLSTM-CNN are shown to have additive and multiplicative cross-structures respectively to deal with the problem. Section~\ref{sec:experiments} evaluates practical effectiveness of the approaches on two challenging NER datasets spanning a wide range of domains with complex, noisy, and emerging entities. The cross-structures bring consistent improvements over the prevalently used Baseline-BiLSTM-CNN without additional gazetteers, POS taggers, language-modeling, or multi-task supervision. The improved core module surpasses comparable bare-bone models on OntoNotes 5.0 and WNUT 2017 by 1.4\% and 4.6\% respectively. Ablation experiments reveal that emerging, complex, confusing, and multi-token entity mentions benefitted much from the cross-structures, up to 8.7\% on some of the multi-token mentions. The in-depth entity-chunking analysis gives insights into how exactly self-attention helps real-world NER. As state-of-the-art approaches often use BiLSTM as their core module, they could benefit from the improvements brought by cross-structures against bare-bone models presented in this paper.

\section{Related Work}
\label{sec:related_work}

Many have attempted tackling the NER task with bare-bone LSTM-based sequence encoders~\citep{Huang:arxiv2015,Ma:ACL2016,Chiu:TACL2016,Lample:NAACL2016}. Among these, the most sophisticated and successful is the BiLSTM-CNN proposed by~\citet{Chiu:TACL2016}. They stack multiple layers of LSTM cells per direction and also use a CNN to compute character-level word vectors alongside pre-trained word vectors. To make the analysis results in this work comparable to past studies on BiLSTM, we largely follow their paper in constructing the Baseline-BiLSTM-CNN, including the selection of raw features, the CNN, and the multi-layer BiLSTM. A subtle difference is that they send the output of each direction through separate affine-softmax classifiers and then sum their probabilities, while this paper sum the scores from affine layers before computing softmax once. While not changing the modeling capacity regarded in this paper, this does provide an empirically stronger baseline model than their formulation.

Besides using additional gazetteers or POS taggers~\citep{Aguilar:WNUT2017,Aguilar:NAACL2018,Ghaddar:COLING2018}, State-of-the-art models use additional large-scale language-modeling corpora~\citep{Akbik:COLING2018} or additional multi-task supervision~\citep{Clark:EMNLP2018} to further improve NER performance beyond bare-bone models. This work does not intend to surpass their performance. Instead, as they rely on a core BiLSTM sentence encoder with the same limitation studied and remedied in this work, they would indeed benefit from the improvements of cross-structures against bare-bone models presented in this paper. In fact, on other tasks, many have used various ways to interleave BiLSTM layers~\citep{Zhou:ACL2015,Coavoux:NAACL2019}. This work provides for a conscious decision with a formal treatment of the XOR limitation and its practical impacts on NER.

The modeling of global contexts for sequence-labeling NER has been partially accomplished using extensive feature engineering or conditional random fields (CRF).~\citet{Ratinov:CoNLL2009} build the Illinois NER tagger with feature-based perceptrons. In their analysis, the usefulness of Viterbi decoding is minimal and conflicts their handcrafted global features. However, their model has limited capability to learn the extraction of new global input features. On the other hand, recent researches on LSTM or CNN-based sequence encoders report empirical improvements brought by CRF~\citep{Huang:arxiv2015,Ma:ACL2016,Lample:NAACL2016,Strubell:EMNLP2017}, as it discourages illegal predictions by explicitly modeling tag-transition probabilities. However, with the speed penalty of Viterbi decoding, transition probabilities are still independent of input sentences and provide partial, limited help in untying two plausible tag sequences. In contrast, this work studies the remedies for the XOR problem of (CRF-)BiLSTM (Section~\ref{sec:xor},~\ref{sec:xor_crf}) that can directly provide the extraction of better global input features, improving class observation likelihoods.

Thought to lighten the burden of compressing all relevant information into a single hidden state, using attention mechanisms on top of LSTMs have shown empirical success for sequence encoders~\citep{Conneau:EMNLP2017,Lin:ICLR2017} and decoders~\citep{Luong:EMNLP2015}. Self-attention has also been used below encoders to compute word vectors conditioned on context~\citep{Devlin:arxiv2018}. This work further formally analyzes the deficiency of BiLSTM encoders for sequence labeling and shows that using self-attention on top is actually providing one type of cross-structures that capture interactions between past and future context.

\section{Model}
\label{sec:model}

\subsection{CNN and Word Features}
\label{sec:cnn}

All models in the experiments use the same set of raw features: character embedding, character type, word embedding, and word capitalization.

For character embedding, 25d vectors are trained from scratch, and 4d one-hot character-type features indicate whether a character is uppercase, lowercase, digit, or punctuation~\citep{Chiu:TACL2016}. Word token lengths are unified to 20 by truncation and padding. The resulting 20-by-(25+4) feature map of each token is applied to a character-trigram CNN with 20 kernels per length 1 to 3 and max-over-time pooling to compute a 60d character-based word vector~\citep{Kim:AAAI2016,Chiu:TACL2016,Ma:ACL2016}.

For word embedding, either pre-trained 300d GloVe vectors~\citep{Pennington:EMNLP2014} or 400d Twitter vectors~\citep{Godin:WNUT2015} are used without further tuning. Also, 4d one-hot word capitalization features indicate whether a word is uppercase, upper-initial, lowercase, or mixed-caps~\citep{Collobert:JMLR2011,Chiu:TACL2016}.

Throughout this paper, $X$ denotes the $n$-by-$d_x$ matrix of sequence features, where $n$ is the sentence length and $d_x$ is either 364 (with GloVe) or 464 (with Twitter).

\subsection{Baseline-BiLSTM-CNN}
\label{sec:baseline}

On top of the feature sequence, BiLSTM is used to capture the future and the past for each time step. Following~\citet{Chiu:TACL2016}, 4 distinct LSTM cells~-- two in each direction~-- are stacked to capture higher level representations:
$$\overrightarrow H=\overrightarrow {LSTM}_2(\overrightarrow {LSTM}_1(X))$$
$$\overleftarrow H=\overleftarrow {LSTM}_4(\overleftarrow {LSTM}_3(X))$$
$$H=\overrightarrow H\ ||\ \overleftarrow H, $$
where $\overrightarrow {LSTM}_i, \overleftarrow {LSTM}_i$ denote applying LSTM cell $i$ in forward, backward order, $\overrightarrow H, \overleftarrow H$ denote the resulting feature matrices of the stacked application, and $||$ denotes row-wise concatenation. In all the experiments, 100d LSTM cells are used, so $H \in R^{n\times d_h}$ and $d_h=200$.

Finally, suppose there are $d_p$ token classes, the probability of each of which is given by the composition of affine and softmax transformations:
$$s_t=H_tW_p+b$$
$$p_{ti}=\frac{e^{s_{ti}}}{\sum_{j=1}^{d_p}e^{s_{tj}}}, $$
where $H_t$ is the $t^{th}$ row of $H$, $W_p\in R^{d_h\times d_p}$, $b\in R^{d_p}$ are a trainable weight matrix and bias, and $s_{ti}$ and $s_{tj}$ are the $i$-th and $j$-th elements of $s_t$.

Following~\citet{Chiu:TACL2016}, the 5 chunk labels \textit{O}, \textit{S}, \textit{B}, \textit{I}, \textit{E} denote if a word token is \textit{O}utside any entity mentions, the \textit{S}ole token of a mention, the \textit{B}eginning token of a multi-token mention, \textit{I}n the middle of a multi-token mention, or the \textit{E}nding token of a multi-token mention. Hence when there are $P$ types of named entities, the actual number of token classes $d_p=P\times 4+1$ for sequence labeling NER.

\subsection{XOR Limitation of Baseline-BiLSTM}
\label{sec:xor}

Consider the following four phrases that form an \textit{XOR}:
\begin{enumerate}
    \itemsep0em 
    \item Key and Peele (\textit{work-of-art})
    \item You and I (\textit{work-of-art})
    \item Key and I
    \item You and Peele
\end{enumerate}
The first two phrases are respectively a show title and a song title. The other two are not entities as a whole, where the last one actually occurs in an interview with Keegan-Michael Key. Suppose each phrase is the sequence given to Baseline-BiLSTM-CNN for sequence tagging, then the $2^{nd}$ token ``and'' should be tagged as \textit{work-of-art:I} in the first two cases and as \textit{O} in the last two cases.

Firstly, note that the score vector at each time step is simply the sum of contributions coming from forward and backward directions plus a bias.
$$s_t=H_tW_p+b$$
$$=\overrightarrow H_t\overrightarrow W_p+\overleftarrow H_t\overleftarrow W_p+b$$
$$=\overrightarrow s_t+\overleftarrow s_t+b$$
where $\overrightarrow W_p,\overleftarrow W_p$ denotes the top-half and bottom-half of $W_p$.

Suppose the index of \textit{work-of-art:I} and \textit{O} are i, j respectively. To predict each ``and'' correctly, it must hold that
$$\overrightarrow s^1_{2i}+\overleftarrow s^1_{2i}+b_i>\overrightarrow s^1_{2j}+\overleftarrow s^1_{2j}+b_j$$
$$\overrightarrow s^2_{2i}+\overleftarrow s^2_{2i}+b_i>\overrightarrow s^2_{2j}+\overleftarrow s^2_{2j}+b_j$$
$$\overrightarrow s^3_{2i}+\overleftarrow s^3_{2i}+b_i<\overrightarrow s^3_{2j}+\overleftarrow s^3_{2j}+b_j$$
$$\overrightarrow s^4_{2i}+\overleftarrow s^4_{2i}+b_i<\overrightarrow s^4_{2j}+\overleftarrow s^4_{2j}+b_j$$
where superscripts denote the phrase number.

Now, the catch is that phrase 1 and phrase 3 have exactly the same past context for ``and''. Hence the same $\overrightarrow H_2$ and the same $\overrightarrow s_2$, i.e., $\overrightarrow s^1_2=\overrightarrow s^3_2$. Similarly, $\overrightarrow s^2_2=\overrightarrow s^4_2$, $\overleftarrow s^1_2=\overleftarrow s^4_2$, and $\overleftarrow s^2_2=\overleftarrow s^3_2$. Rewriting the constraints with these equalities gives
$$\overrightarrow s^1_{2i}+\overleftarrow s^1_{2i}+b_i>\overrightarrow s^1_{2j}+\overleftarrow s^1_{2j}+b_j$$
$$\overrightarrow s^2_{2i}+\overleftarrow s^2_{2i}+b_i>\overrightarrow s^2_{2j}+\overleftarrow s^2_{2j}+b_j$$
$$\overrightarrow s^1_{2i}+\overleftarrow s^2_{2i}+b_i<\overrightarrow s^1_{2j}+\overleftarrow s^2_{2j}+b_j$$
$$\overrightarrow s^2_{2i}+\overleftarrow s^1_{2i}+b_i<\overrightarrow s^2_{2j}+\overleftarrow s^1_{2j}+b_j$$

Finally, summing the first two inequalities and the last two inequalities gives two contradicting constraints that cannot be satisfied. In other words, even if an oracle is given to training the model, Baseline-BiLSTM-CNN can only tag at most 3 out of 4 ``and'' correctly. No matter how many LSTM cells are stacked for each direction, the formulation in previous studies simply does not have enough modeling capacity to capture cross-context patterns for sequence labeling NER.

\begin{table*}
\caption{
    \label{tab:results:overall} Overall results.
    *Used on WNUT for character-based word vectors, reported better than CNN.
}
\centering
\begin{tabular}{|l|ccc|ccc|}
\hline\Tstrut
 & \multicolumn{3}{c|}{OntoNotes 5.0} & \multicolumn{3}{c|}{WNUT 2017} \\
 & Prec. & Rec. & F1 & Prec. & Rec. & F1 \\
\hline\Tstrut
BiLSTM-CNN           & 86.04 & 86.53 & 86.28$\pm$0.26 & - & - &     - \\
CRF-IDCNN            &     - &     - & 86.84$\pm$0.19 & - & - &     - \\
CRF-BiLSTM(-BiLSTM*) &     - &     - & 86.99$\pm$0.22 & - & - & 38.24 \\
\hline\Tstrut
Baseline-BiLSTM-CNN &          88.37 &          87.14 &          87.75$\pm$0.14 &          53.24 &          32.93 &          40.68$\pm$1.78 \\
Cross-BiLSTM-CNN    &          88.37 & \textbf{88.17} &          88.27$\pm$0.17 & \textbf{58.28} &          33.92 & \textbf{42.85}$\pm$0.99 \\
Att-BiLSTM-CNN      & \textbf{88.71} &          88.11 & \textbf{88.40}$\pm$0.18 &          55.82 & \textbf{34.08} &          42.26$\pm$0.82 \\
\hline
\end{tabular}
\end{table*}

\begin{table}
\caption{\label{tab:dataset:statistics} Datasets (K-tokens / K-entities).}
\centering
\begin{tabular}{|r|r|c|}
\hline
 & OntoNotes 5.0 & WNUT 2017 \\
\hline\Tstrut
train & 1088.5 / 81.8 & 62.7 / 1.9 \\
dev   &  147.7 / 11.0 & 15.7 / 0.8 \\
test  &  152.7 / 11.2 & 23.3 / 1.0 \\
\hline
\end{tabular}
\end{table}

\subsection{XOR Limitation of CRF-BiLSTM}
\label{sec:xor_crf}

Consider the following four phrases that form an \textit{XOR}:
$$a\_o\ m\_s\ c\_o$$
$$b\_o\ m\_s\ d\_o$$
$$a\_o\ m\_o\ d\_o$$
$$b\_o\ m\_o\ c\_o$$
a, b, m, c, d denote words. \textit{s} (single) and \textit{o} (outside) are tags.

The correct tagging of all phrases requires that
$$p(oso|amc) > p(ooo|amc)$$
$$p(oso|bmd) > p(ooo|bmd)$$
$$p(oso|amd) < p(ooo|amd)$$
$$p(oso|bmc) < p(ooo|bmc)$$
Note that this time we consider each phrase as a whole.

Suppose there is only Softmax, the log-probability of a phrase is just the log-sum of each time step. Cancelling the same terms across two sides of each inequality, e.g. $lp(o\_\_|amc)$, gives
$$lp(\_s\_|amc) > lp(\_o\_|amc)$$
$$lp(\_s\_|bmd) > lp(\_o\_|bmd)$$
$$lp(\_s\_|amd) < lp(\_o\_|amd)$$
$$lp(\_s\_|bmc) < lp(\_o\_|bmc)$$

Without cross-structures, scores from two contexts are only linearly summed (See Section~\ref{sec:xor}), which gives
$$lp(\_s|am) + lp(s\_|mc) > lp(\_o|am) + lp(o\_|mc)$$
$$lp(\_s|bm) + lp(s\_|md) > lp(\_o|bm) + lp(o\_|md)$$
$$lp(\_s|am) + lp(s\_|md) < lp(\_o|am) + lp(o\_|md)$$
$$lp(\_s|bm) + lp(s\_|mc) < lp(\_o|bm) + lp(o\_|mc)$$
For pure BiLSTM, the original proof sums the top 2 and the bottom 2 inequalities, resulting in contradicting constraints.

Now, if there had been a linear-chain CRF modeling label transition probabilities (call it $q$), it would only add yet another linear term and would require \\
$lp(\_s|am) + lp(s\_|mc) + lq(oso) >$ \begin{flushright}$lp(\_o|am) + lp(o\_|mc) + lq(ooo)$\end{flushright}
$lp(\_s|bm) + lp(s\_|md) + lq(oso) >$ \begin{flushright}$lp(\_o|bm) + lp(o\_|md) + lq(ooo)$\end{flushright}
$lp(\_s|am) + lp(s\_|md) + lq(oso) <$ \begin{flushright}$lp(\_o|am) + lp(o\_|md) + lq(ooo)$\end{flushright}
$lp(\_s|bm) + lp(s\_|mc) + lq(oso) <$ \begin{flushright}$lp(\_o|bm) + lp(o\_|mc) + lq(ooo)$\end{flushright}
The inequalities remain unsatisfiable, and the reason is two-fold:
\begin{enumerate}
    \itemsep0em 
    \item The addition of transition probabilities are linear, independent of word sequences, so it does not help untying plausible word-tag sequences that form XOR.
    \item The consideration of each phrase as a whole, i.e. Viterbi decoding, does help to untie \textit{BIE} with \textit{OOO}, but not to untie \textit{OSO} with \textit{OOO} (recall ``cancelling the same terms'').
\end{enumerate}
In other words, predicting a phrase as a whole partially mitigates the XOR problem, with or without transition probabilities.

\subsection{Cross-BiLSTM-CNN}
\label{sec:cross}

Motivated by the limitation of the conventional Baseline-BiLSTM-CNN for sequence labeling, this paper proposes the use of Cross-BiLSTM-CNN by changing the deep structure in Section~\ref{sec:baseline} to
$$\overrightarrow H^1=\overrightarrow {LSTM}_1(X)$$
$$\overleftarrow H^3=\overleftarrow {LSTM}_3(X)$$
$$\overrightarrow H^2=\overrightarrow {LSTM}_2(\overrightarrow H^1||\overleftarrow H^3)$$
$$\overleftarrow H^4=\overleftarrow {LSTM}_4(\overrightarrow H^1||\overleftarrow H^3)$$
$$H=\overrightarrow H^2\ ||\ \overleftarrow H^4$$

As the forward and backward hidden states are interleaved between stacked LSTM layers, Cross-BiLSTM-CNN models cross-context patterns by computing representations of the whole sequence in a feed-forward, additive manner.

Specifically, for the XOR cases introduced in Section~\ref{sec:xor},~\ref{sec:xor_crf}, although phrase 1 and phrase 3 still have the same past context for the middle token and hence the first layer $\overrightarrow {LSTM}_1$ can only extract the same low-level hidden features $\overrightarrow H^1_2$, the second layer $\overrightarrow {LSTM}_2$ considers the whole context $\overrightarrow H^1||\overleftarrow H^3$ and thus have the ability to extract different high-level hidden features $\overrightarrow H^2_2$ for the two phrases.

As the higher-level LSTMs of Cross-BiLSTM-CNN have interleaved input from forward and backward hidden states down below, their weight parameters double the size of the first-level LSTMs. Nevertheless, the cross formulation provides the modeling capacity absent in previous studies with how many more LSTM layers.

\begin{table*}
\caption{
    \label{tab:results:entitytype} Types with significant results ($>$3\% absolute F1 differences vs. Baseline); .
    *Nationalities, religious, political groups.
}
\centering
\begin{tabular}{|l|ccccc|ccc|}
\hline\Tstrut
 & \multicolumn{5}{c|}{OntoNotes 5.0} & \multicolumn{3}{c|}{WNUT 2017} \\
 & event & language & law & NORP* & work-of-art & corporation & creative-work & location \\
\hline\Tstrut
Cross & +3.0\% & +4.1\% & +4.5\% & +3.3\% & +2.1\% & +6.4\% & +3.2\% & +8.6\% \\
Att   & +4.6\% & +0.8\% & +0.8\% & +3.4\% & +5.6\% & +0.3\% & +2.0\% & +5.3\% \\
\hline
\end{tabular}
\end{table*}

\begin{table*}
\caption{\label{tab:results:entitylength} Improvements vs. Baseline among different mention lengths.}
\centering
\begin{tabular}{|l|cccc|cccc|}
\hline\Tstrut
 & \multicolumn{4}{c|}{OntoNotes 5.0} & \multicolumn{4}{c|}{WNUT 2017} \\
 & 1 & 2 & 3 & 3+ & 1 & 2 & 3 & 3+ \\
\hline\Tstrut
Cross & +0.3\% & +0.6\% & +1.8\% & +1.3\% & +1.7\% & +2.9\% & +8.7\% & +5.4\% \\
Att   & +0.1\% & +1.1\% & +2.3\% & +1.8\% & +1.5\% & +2.0\% & +2.6\% & +0.9\% \\
\hline
\end{tabular}
\end{table*}

\subsection{Att-BiLSTM-CNN}
\label{sec:att}

Another way to capture the interaction between past and future context per time step is to add a token-level self-attentive mechanism on top of the same BiLSTM formulation introduced in Section~\ref{sec:baseline}. Given the hidden features $H$ of a whole sequence, the model projects each hidden state to different subspaces, depending on whether it is used as the \textit{q}uery vector to consult other hidden states for each word token, the \textit{k}ey vector to compute its dot-similarities with incoming queries, or the \textit{v}alue vector to be weighted and actually convey information to the querying token. As different aspects of a task can call for different attention, multiple attention heads running in parallel are used~\citep{Vaswani:NIPS2017}.

Formally, let $m$ be the number of attention heads and $d_c$ be the subspace dimension. For each head $i\in \{1..m\}$, the attention weight matrix and context matrix are computed by
$$\alpha^i=\sigma \left(\frac{HW^{qi}{(HW^{ki})}^T}{\sqrt{d_c}}\right)$$
$$C^i=\alpha^iHW^{vi}, $$
where $W^{qi},W^{ki},W^{vi}\in R^{d_h\times d_c}$ are trainable projection matrices and $\sigma$ performs softmax along the second dimension. Each row of the resulting $\alpha^1,\alpha^2,\ldots,\alpha^m\in R^{n\times n}$ contains the attention weights of a token to its context, and each row of $C^1,C^2,\ldots,C^m\in R^{n\times d_c}$ is its context vector.

For Att-BiLSTM-CNN, the hidden vector and context vectors of each token are considered together for classification:
$$s_t^c=(H_t||C^1_t||C^2_t||...||C^m_t)W_c+b$$
$$p_{ti}^c=\frac{e^{s_{ti}^c}}{\sum_{j=1}^{d_p}e^{s_{tj}^c}}, $$
where $C^i_t$ is the $t$-th row of $C^i$, and $W_c\in R^{(d_h+md_c)\times d_p}$ is a trainable weight matrix. In all the experiments, $m=5$ and $d_c=\frac{d_h}{5}$, so $W_c\in R^{2d_h\times d_p}$.

While the BiLSTM formulation stays the same as Baseline-BiLSTM-CNN, the computation of attention weights $\alpha^i$ and context features $C^i$ models the cross interaction between past and future. To see this, the computation of attention scores can be rewritten as follows.
$$HW^{qi}{(HW^{ki})}^T=H(W^{qi}{W^{ki}}^T)H^T.$$
$$=(\overrightarrow H\ ||\ \overleftarrow H)(W^{qi}{W^{ki}}^T){(\overrightarrow H\ ||\ \overleftarrow H)}^T.$$
With the un-shifted covariance matrix of the projected $\overrightarrow H\ ||\ \overleftarrow H$, Att-BiLSTM-CNN correlates past and future context for each token in a dot-product, multiplicative manner.

One advantage of using multiplicative attention to resolve the XOR problem is that it only needs to be computed once per sequence, and the matrix computations are highly parallelizable, resulting in little computation time overhead. Moreover, in Section~\ref{sec:experiments}, the attention weights provide a better understanding of how the model learns to tackle sequence-labeling NER.

\section{Experiments}
\label{sec:experiments}

\subsection{Datasets}
\label{sec:datasets}

\textbf{OntoNotes 5.0 Fine-Grained NER}~-- a million-token corpus with diverse sources of newswires, web, broadcast news, broadcast conversations, magazines, and telephone conversations~\citep{Hovy:NAACL2006,Pradhan:CoNLL2013}. Some are transcriptions of talk shows, and some are translations from Chinese or Arabic. The dataset contains 18 fine-grained entity types, including hard ones such as \textit{law}, \textit{event}, and \textit{work-of-art}. All the diversities and noisiness require that models are robust across broad domains and able to capture a multitude of linguistic patterns for complex entities.

\textbf{WNUT 2017 Emerging NER}~-- a dataset providing maximally diverse, noisy, and drifting user-generated text~\cite{Derczynski:WNUT2017}. The training set consists of previously annotated tweets~-- social media text with non-standard spellings, abbreviations, and unreliable capitalization~\citep{Strauss:WNUT2016}; the development set consists of newly sampled YouTube comments; the test set includes text newly drawn from Twitter, Reddit, and StackExchange. Besides drawing new samples from diverse topics across different sources, the shared task also filtered out text containing surface forms of entities seen in the training set. The resulting dataset requires models to generalize to emerging contexts and entities instead of relying on familiar surface cues.

\begin{figure*}
    \centering
    
    \begin{subfigure}{1.5\columnwidth}
        \centering
        \includegraphics[width=\columnwidth]{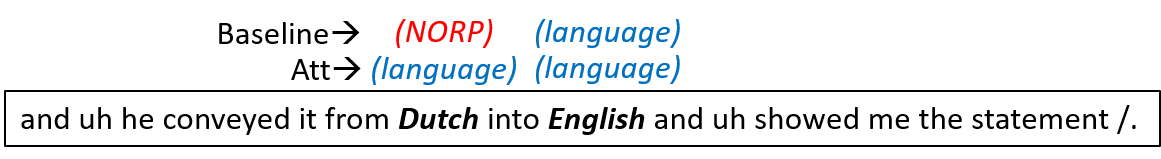}
        \caption{\label{fig:dutch:sentence}A confusing surface form for \textit{language} and \textit{nationality}.}
    \end{subfigure}\\
    \par\smallskip
    \begin{subfigure}{1.95\columnwidth}
        \centering
        \includegraphics[width=\columnwidth]{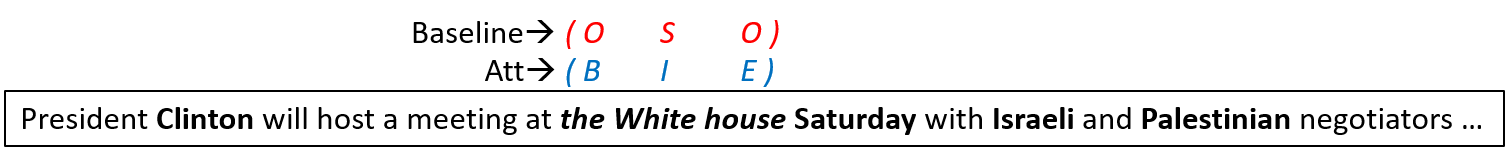}
        \caption{\label{fig:white:sentence}A triple-token mention with unreliable capitalization.}
    \end{subfigure}\\
    
    \caption{Example problematic entities for Baseline-BiLSTM-CNN.}
    \label{fig:case:sentence}
\end{figure*}

\begin{table*}
\caption{\label{tab:results:chunking} Entity-chunking ablation results.}
\centering
\begin{tabular}{|c|r|r|r|r|r|r|r|r|r|}
\hline\Tstrut
 & \multicolumn{8}{c|}{Att-BiLSTM-CNN} & Baseline-... \\
\cline{3-9}\Tstrut 
 & $HC^{all}$ & $H$ & $C^{all}$ & $C^1$ & $C^2$ & $C^3$ & $C^4$ & $C^5$ & $H$ \\
O & 99.05 &  -1.68 &   0.75 &   0.95 &  -1.67 & -45.57 &  -0.81 & -35.46 &  -0.03 \\
S & 93.74 &   2.69 & -91.02 & -90.56 & -90.88 & -25.61 & -86.25 & -84.32 &   \bb{0.13} \\
B & 90.99 &   1.21 & -52.26 & -90.78 & -88.08 & -90.88 & \bb{-12.21} & -87.45 &  \br{-0.63} \\
I & 90.09 & \br{-28.18} &  \bb{-3.80} & -87.93 & \bb{-60.56} & \bb{-50.19} & \bb{-57.19} & -79.63 &  \br{-0.41} \\
E & 93.23 &   2.00 & -71.50 & -93.12 & \bb{-36.45} & \bb{-39.19} & -91.90 & -90.83 &  \br{-0.38} \\
\hline
\end{tabular}
\end{table*}

\subsection{Implementation and Baselines}
\label{sec:implementation}

All experiments for Baseline-, Cross-, and Att-BiLSTM-CNN used the same model parameters given in Section~\ref{sec:model}. The training minimized per-token cross-entropy loss with the Nadam optimizer~\citep{Dozat:ICLR2016} with uniform learning rate 0.001, batch size 32, and 35\% variational dropout~\citep{Gal:NIPS2016}. Each training lasted 400 epochs when using GloVe embedding (OntoNotes), and 1600 epochs when using Twitter embedding (WNUT). The development set of each dataset was used to select the best epoch to restore model weights for testing. Following previous work on NER, model performances were evaluated with strict mention F1 score. Training of each model on each dataset repeated 6 times to report the mean score and standard deviation.

Besides the strong Baseline implemented in this paper, we also list results of bare-bone BiLSTM-CNN~\citep{Chiu:TACL2016}, CRF-BiLSTM(-BiLSTM)~\citep{Strubell:EMNLP2017,Lin:WNUT2017}, and CRF-IDCNN~\citep{Strubell:EMNLP2017} from the literature. Among them, IDCNN was a CNN-based sentence encoder, which should not have the XOR limitation raised in this paper. \textbf{Caveat}: As the purpose of the experiments is to evaluate practical effectiveness in remedying the limitation of BiLSTM, comparisons are not made against models using additional resources, such as gazetteers or POS taggers~\citep{Aguilar:WNUT2017,Aguilar:NAACL2018,Ghaddar:COLING2018}, large-scale language-modeling corpora~\citep{Akbik:COLING2018}, or multi-task supervision~\citep{Clark:EMNLP2018}, to further improve NER performance beyond bare-bone models. We do not claim to have surpassed state-of-the-art results. However, as they used BiLSTM sentence encoders with the XOR limitation, they could indeed integrate with and benefit from the cross-structures presented in this paper.

\subsection{Overall Results}
\label{sec:overall_results}

Table~\ref{tab:results:overall} shows overall results on the two datasets spanning broad domains of newswires, broadcast, telephone, and social media. The models proposed in this paper surpassed previous reported bare-bone models by 1.4\% on OntoNotes and 4.6\% on WNUT. Compared to the re-implemented Baseline-BiLSTM-CNN, the cross-structures brought 0.7\% and 2.2\% improvements on OntoNotes and WNUT. More substantial improvements were achieved for WNUT 2017 emerging NER, suggesting that cross-context patterns were even more crucial for emerging contexts and entities than familiar entities, which might often be memorized by their surface forms.

\subsection{Complex and Confusing Entity Mentions}
\label{sec:complex_and_confusing}

Table~\ref{tab:results:entitytype} shows significant results per entity type compared to Baseline ($>$3\% absolute F1 differences for either Cross or Att). It could be seen that harder entity types generally benefitted more from the cross-structures. For example, \textit{work-of-art/creative-work} entities could in principle take any surface forms~-- unseen, the same as a person name, abbreviated, or written with unreliable capitalizations on social media. Such mentions require models to learn a deep, generalized understanding of their context to accurately identify their boundaries and disambiguate their types. Both cross-structures were more capable in dealing with such hard entities (2.1\%/5.6\%/3.2\%/2.0\%) than the prevalently used, problematic Baseline.

\begin{figure*}
    \centering
    
    \begin{subfigure}{1.95\columnwidth}
        \centering
        \includegraphics[width=\columnwidth]{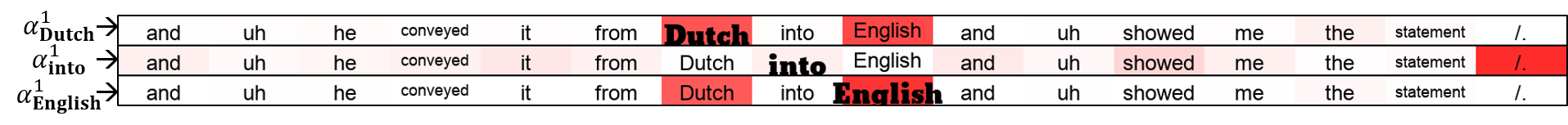}
        \caption{\label{fig:dutch:head1}(Partial) $\alpha^1$ of "...Dutch into English...".}
    \end{subfigure}\\
    \par\bigskip
    \begin{subfigure}{1.95\columnwidth}
        \centering
        \includegraphics[width=\columnwidth]{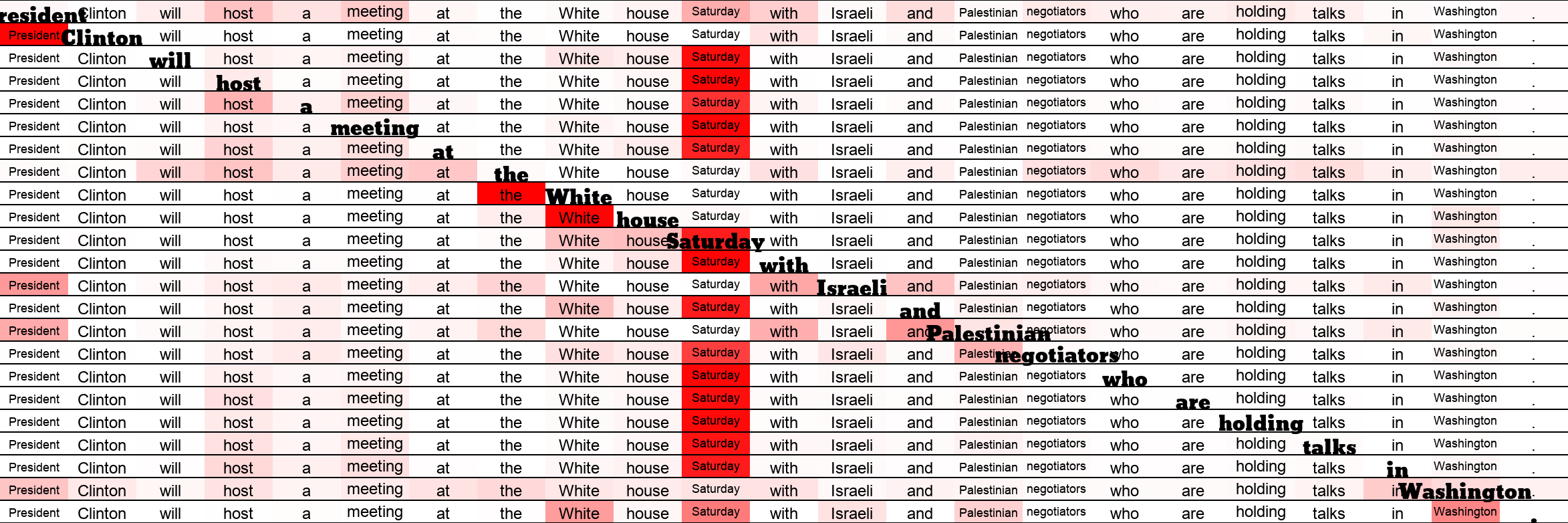}
        \caption{\label{fig:white:head2}$\alpha^2$ of "...the White house...".}
    \end{subfigure}\\
    \par\bigskip
    \begin{subfigure}{1.95\columnwidth}
        \centering
        \includegraphics[width=\columnwidth]{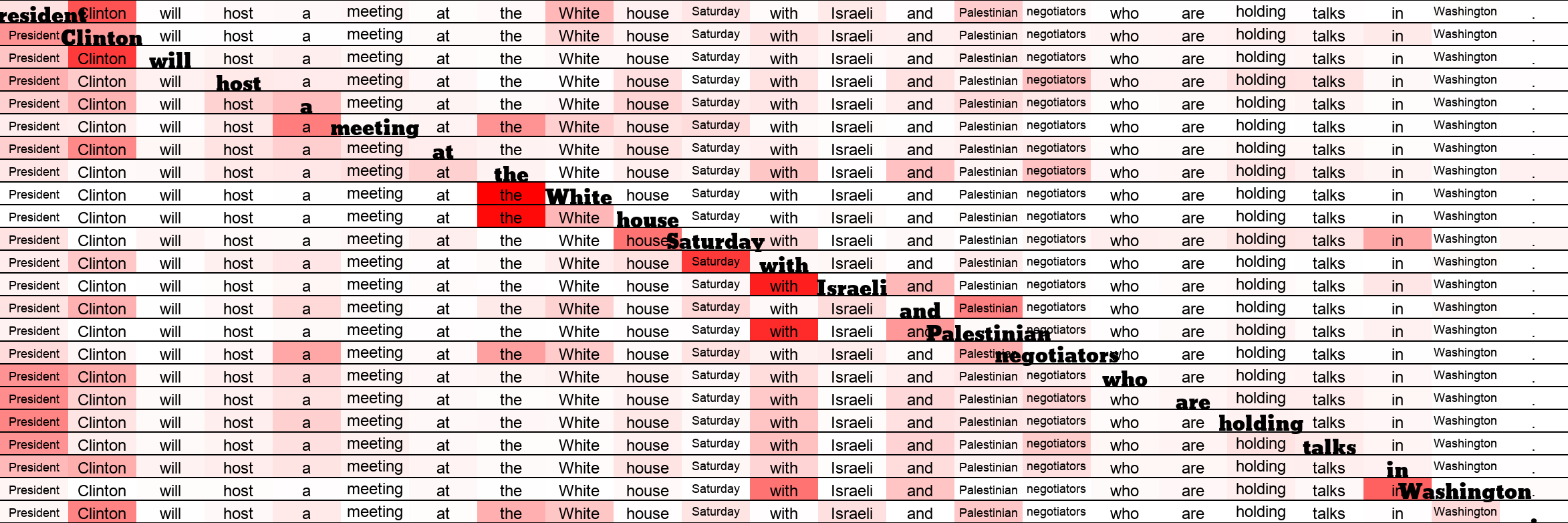}
        \caption{\label{fig:white:head3}$\alpha^3$ of "...the White house...".}
    \end{subfigure}\\
    \par\bigskip
    \begin{subfigure}{1.95\columnwidth}
        \centering
        \includegraphics[width=\columnwidth]{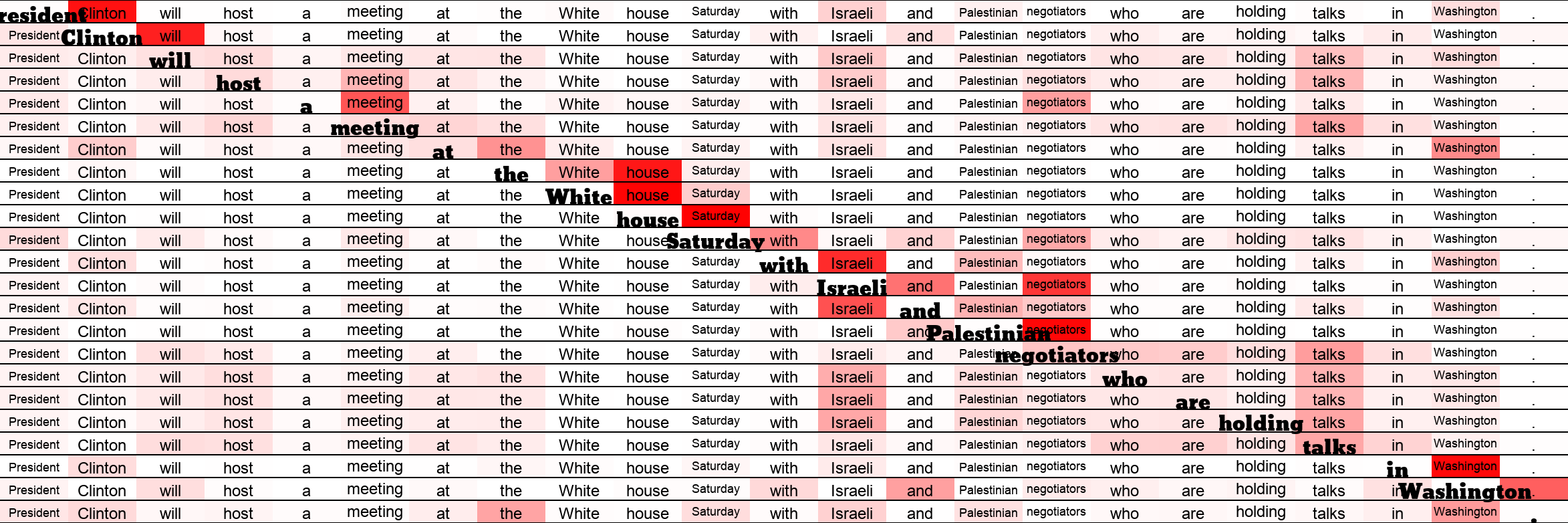}
        \caption{\label{fig:white:head4}$\alpha^4$ of "...the White house...".}
    \end{subfigure}\\
    
    \caption{Attention heat maps for the mentions in Figure~\ref{fig:case:sentence}, best viewed on computer.}
    \label{fig:case:head}
\end{figure*}

Moreover, disambiguating fine-grained entity types is also a challenging task. For example, entities of \textit{language} and \textit{NORP} often take the same surface forms. Figure~\ref{fig:dutch:sentence} shows an example containing "Dutch" and "English". While "English" was much more frequently used as a \textit{language} and was identified correctly, the "Dutch" mention was tricky for Baseline. The attention heat map (Figure~\ref{fig:dutch:head1}) further tells the story that Att has relied on its attention head to make context-aware decisions. Overall, both cross-structures were much better at disambiguating these fine-grained types (4.1\%/0.8\%/3.3\%/3.4\%).

\subsection{Multi-Token Entity Mentions}
\label{sec:multi_token}

Table~\ref{tab:results:entitylength} shows results among different entity lengths. It could be seen that cross-structures were much better at dealing with multi-token mentions compared to the prevalently used, problematic Baseline.

In fact, identifying correct mention boundaries for multi-token mentions poses a unique challenge for sequence-labeling models~-- all tokens in a mention must be tagged with correct sequential labels to form one correct prediction. Although models often rely on strong hints from a token itself or a single side of the context, however, in general, cross-context modeling is required. For example, a token should be tagged as \textit{I}nside if and only if it immediately follows a \textit{B}egin or an \textit{I} and is immediately followed by an \textit{I} or an \textit{E}nd.

Figure~\ref{fig:white:sentence} shows a sentence with multiple entity mentions. Among them, "the White house" is a triple-token facility mention with unreliable capitalization, resulting in an emerging surface form. Without usual strong hints given by a seen surface form, Baseline predicted a false single-token mention "White". In contrast, Att utilized its multiple attention heads (Figure~\ref{fig:white:head2},~\ref{fig:white:head3},~\ref{fig:white:head4}) to consider the preceding and succeeding tokens for each token and correctly tagged the three tokens as \textit{facility:B}, \textit{facility:I}, \textit{facility:E}.

\subsection{Entity-Chunking}
\label{sec:entity_chunking}

Entity-chunking is a subtask of NER concerned with locating entity mentions and their boundaries without disambiguating their types. For sequence-labeling models, this means correct \textit{O}, \textit{S}, \textit{B}, \textit{I}, \textit{E} tagging for each token. In addition to showing that cross-structures achieved superior performance on multi-token entity mentions (Section~\ref{sec:multi_token}), an ablation study focused on the chunking tags was performed to better understand how it was achieved.

Table~\ref{tab:results:chunking} shows the entity-chunking ablation results on OntoNotes 5.0 development set. Both Att and Baseline models were taken without re-training for this subtask. The $HC^{all}$ column lists the performance of Att-BiLSTM-CNN on each chunking tag. Other columns list the performance compared to $HC^{all}$. Columns $H$ to $C^5$ are when the full model is deprived of all other information in testing time by forcefully zeroing all vectors except the one specified by the column header. The figures shown in the table are per-token recalls for each chunking tag, which tells if a part of the model is responsible for signaling the whole model to predict that tag. Bold font and underline mark relatively \bb{high} and \br{low} values of interest.

Firstly, Att appeared to designate the task of scoring \textit{I} to the attention mechanism: When context vectors $C^{all}$ were left alone, the recall for \textit{I} tokens only dropped a little (\bb{-3.80}); When token hidden states $H$ were left alone, the recall for \textit{I} tokens seriously degraded (\br{-28.18}). When $H$ and $C^{all}$ work together, the full Att model was then better at predicting multi-token entity mentions than Baseline.

Then, breaking context vectors to each attention head reveals that they have worked in cooperation: $C^2$, $C^3$ focused more on scoring \textit{E} (\bb{-36.45}, \bb{-39.19}) than \textit{I} (\bb{-60.56, -50.19}), while $C^4$ focused more on scoring \textit{B} (\bb{-12.21}) than \textit{I} (\bb{-57.19}). It was when information from all these heads were combined was Att able to better identify a token as being \textit{I}nside a multi-token mention than Baseline.

Finally, the quantitative ablation analysis of chunking tags in this Section and the qualitative case-study attention visualizations in Section~\ref{sec:multi_token} explains each other: $C^2$ and especially $C^3$ tended to focus on \textbf{looking for immediate preceding mention tokens} (the diagonal shifted left in Figure~\ref{fig:white:head2},~\ref{fig:white:head3}), enabling them \textbf{to signal for \textit{E}nd and \textit{I}nside}; $C^4$ tended to focus on \textbf{looking for immediate succeeding mention tokens} (the diagonal shifted right in Figure~\ref{fig:white:head4}), enabling it \textbf{to signal for \textit{B}egin and \textit{I}nside}. In fact, without context vectors, instead of \textit{BIE}, Att would tag "the White house" as \textit{BSE} and extract the same false mention of "White" as the \textit{OSO} of Baseline.

Lacking the ability to model cross-context patterns, Baseline inadvertently learned to retract to predict single-token entities (\bb{0.13} vs. \br{-0.63, -0.41, -0.38}) when an easy hint from a familiar surface form is not available. This indicates a major flaw in BiLSTM-CNNs prevalently used for real-world NER today.

\section{Conclusion}
\label{sec:conclusion}

This paper has given a formal treatment of the deficiency of the prevalently-used (CRF-)BiLSTM-CNN in modeling cross-context for sequence-labeling NER. Formal proof of its inability to capture XOR patterns has been given, and the practical impacts has been analyzed on OntoNotes 5.0 and WNUT 2017. Additive and multiplicative cross-structures have shown to be crucial in modeling cross-context, significantly enhancing recognition of emerging, complex, confusing, and multi-token entity mentions. Against comparable bare-bone models, 1.4\% and 4.6\% overall improvements on OntoNotes 5.0 and WNUT 2017 have been achieved, showing the importance of remedying the core module of NER. As state-of-the-art models use (CRF-)BiLSTM with XOR limitation, this study should lay a sound foundation for future improvements on sequence-labeling NER.

\bibliographystyle{aaai}
\bibliography{AAAI-LiP.6792}

\end{document}